# LPRnet: A self-supervised registration network for LiDAR and photogrammetric point clouds

Chen Wang, *Student Member, IEEE,* Yanfeng Gu, *Senior Member, IEEE* and Xian Li, *Member, IEEE*

*Abstract*—LiDAR and photogrammetry are active and passive remote sensing techniques for point cloud acquisition, respectively, offering complementary advantages and heterogeneous. Due to the fundamental differences in sensing mechanisms, spatial distributions and coordinate systems, their point clouds exhibit significant discrepancies in density, precision, noise, and overlap. Coupled with the lack of ground truth for large-scale scenes, integrating the heterogeneous point clouds is a highly challenging task. This paper proposes a self-supervised registration network based on a masked autoencoder, focusing on heterogeneous LiDAR and photogrammetric point clouds. At its core, the method introduces a multi-scale masked training strategy to extract robust features from heterogeneous point clouds under self-supervision. To further enhance registration performance, a rotation-translation embedding module is designed to effectively capture the key features essential for accurate rigid transformations. Building upon the robust representations, a transformer-based architecture seamlessly integrates local and global features, fostering precise alignment across diverse point cloud datasets. The proposed method demonstrates strong feature extraction capabilities for both LiDAR and photogrammetric point clouds, addressing the challenges of acquiring ground truth at the scene level. Experiments conducted on two real-world datasets validate the effectiveness of the proposed method in solving heterogeneous point cloud registration problems.

## I. Introduction

Point cloud serves as a data format for representing three-dimensional (3D) data by recording the coordinates and intensity values of points on the surfaces of real-world objects. Due to the capacity to capture 3D information, point cloud has attracted considerable research interest in recent years, playing crucial roles in fields such as 3D mapping and autonomous driving [1-3]. In remote sensing applications, point cloud addresses challenges associated with classification and target detection in two-dimensional images such as RGB, multispectral, and hyperspectral images by incorporating elevation data [4-7].

Point cloud can be acquired through various techniques including LiDAR and photogrammetry, each with distinct strengths and limitations [8-10]. LiDAR systems actively generate highly accurate 3D spatial data with centimeter-level precision and penetration capability, making them ideal for applications like terrain mapping and forestry. However, high cost and lack of spectral information limit their broader applications. Photogrammetric point clouds, passively derived from overlapping optical images, provide dense 3D reconstructions with rich texture and spectral information. Despite the cost-effectiveness, photogrammetric point clouds suffer from lower geometric accuracy and vulnerability to occlusion. Given the complementary strengths of LiDAR and photogrammetric point clouds, integrating these heterogeneous data via registration becomes essential to leverage their respective advantages.

Heterogeneous point cloud registration involves aligning point clouds from different sources, such as LiDAR and photogrammetry, into a common coordinate system [11, 12]. Unlike homogeneous registration [13], where the data share similar characteristics, heterogeneous registration faces unique challenges due to the inherent discrepancies in density, precision, noise, and overlap [14, 15]. The lack of common features further complicates the alignment process. The detailed differences and principles will be explained in Section II. These challenges necessitate robust registration frameworks capable of addressing discrepancies in spatial distribution and feature representation. Existing heterogeneous point cloud registration methods can be mainly divided into three categories: optimization-based, feature-based, and end-to-end learning-based.

The optimization-based method focuses on establishing correspondences between the reference point cloud and the point cloud to be registered. The most representative algorithms include the Iterative Closest Point (ICP) method and its derivatives [16-18]. The original ICP algorithm utilizes the distance between points as the similarity measure for the correspondence relationship, constructs an objective function that quantifies the error between point clouds using the least squares method, and solves for the rigid transformation matrix that minimizes this objective function through iterative optimization. Derivative algorithms, such as Fast and Robust ICP and globally optimal algorithm ICP (GO ICP), enhance point cloud similarity measures and optimization algorithms to improve registration efficiency [17, 18]. Although these methods have achieved good results in homogeneous point cloud registration, heterogeneous registration presents additional challenges due to more pronounced differences in

Manuscript received xx xx, 2024; revised Xxx xx, 2024; accepted Xxx xx, 2024. This work was supported by the National Science Fund for Distinguished Young Scholars under Grant 62025107 and the Major Scientific Instrument Development Program of the National Natural Science Foundation of China under Grant 62327803, and in part by the Open Fund Project of KuiYuan Laboratory (Grant No. KY202423) (*Corresponding author: Xian Li.*)

The authors are with the School of Electronics and Information Engineering, Harbin Institute of Technology, Harbin 150001, China (e-mail: xianli@hit.edu.cn).

point cloud characteristics. Probabilistic methods are another branch of optimization methods. For instance, the GMM method models the point cloud as a Gaussian mixture model and uses the expectation-maximization algorithm to determine the optimal transformation [19]. Huang et al. proposed a two-stage registration algorithm, which combines shape functions descriptor and generative Gaussian mixture model to address the noise issues in heterogeneous point clouds [20].

Feature-based methods first extract feature points and their descriptors from point cloud data, subsequently estimating the transformation between two point clouds by matching these feature points without the need for iterative optimization. Features can include both handcrafted features, such as Simplified Point Feature Histograms (SPFH) and Fast Point Feature Histograms (FPFH) [21, 22], and those derived from deep learning techniques [23-25]. By developing network architectures and designing appropriate loss functions, salient features can be learned from point clouds to achieve robust mapping of spatial relationships. FCGF obtains point cloud features through a fully convolutional network and captures a wide range of spatial context information [26]. PPFnet uses the Pointnet structure to learn globally aware 3D descriptors. These feature extraction networks can effectively capture robust spatial features from point clouds [27]. However, feature-based methods typically require voxelization of unordered point clouds, as seen in FCGF and SpinNet methods [28]. Applying a unified voxelization scale across heterogeneous point clouds is often impractical. Similarly, point-based feature extraction methods, such as D3Feat [29], require selecting the number of points in a neighborhood, which can be heavily influenced by point cloud density. More complex models, such as multi-resolution and robust feature-based methods, are required to handle the effects of heterogeneous point cloud density [30]. Zhao et al. proposed a spherical voxel feature representation method, which achieved density-consistent feature representation and avoided feature loss caused by down sampling [31]. Sufficient training data enables such methods to learn the correspondence between feature points [32].

Inspired by self-supervised models for images and languages, a series of self-supervised training approaches for point clouds have been developed, achieving notable success in downstream tasks such as object classification and part segmentation. For instance, PointBERT leverages DGCNN to tokenize point clouds and employs contrastive learning to extract high-level semantic features [33]. Similarly, PointMAE introduces the Masked Autoencoder framework to the point cloud domain, utilizing a standard transformer for pretraining on point cloud reconstruction [34]. However, these methods do not extend their applicability to downstream tasks like point cloud registration.

End-to-end learning-based methods take both the reference and target point clouds as inputs to a neural network, directly learning the transformation relationship through the network's regression capabilities. For instance, DeepVCP improves registration accuracy by establishing virtual corresponding points and leveraging the matching probability between candidate points [35]. PointNetLK applies the PointNet architecture to global point cloud registration, introducing an approach that eliminates the need for explicit correspondences [36, 37]. Unlike feature-based methods, the end-to-end approach not only emphasizes the extraction of point cloud features but also incorporates transformation estimation information within the network. End-to-end deep learning methods often depend on ground truth transformation values between point clouds, which are relatively easy to obtain in indoor or small-scale scenes. However, for large-scale remote sensing data, particularly heterogeneous data, acquiring such ground truth data is challenging. To expand the scope of application, FMR method aligns two point clouds by minimizing the global feature projection error without ground truth, but ignores local information and distribution differences of heterogeneous point clouds [38].

To address the challenges outlined above, we propose a self-supervised point cloud registration method designed for irregular and low-overlap LiDAR and photogrammetric point clouds. The network introduces a masked autoencoder-based architecture, offering robustness to local incompleteness in heterogeneous point clouds. The masked autoencoder is composed of three key components: a multi-scale random masking module simulates the characteristics of heterogeneous data, a registration feature embedding module integrates local rotational and translational features, and a transformer module that unifies global feature representations. During the training phase, the input point clouds are randomly masked, and a reconstruction loss function is employed to enable self-supervised learning. In the testing phase, the method combines deep learning-derived features with optimization algorithms by minimizing the feature distance between the target and reference point clouds to solve for the transformation matrix. To assess the effectiveness of the proposed registration method, we tested it on two sets of real-world point clouds. Experimental results from the test areas indicate that our method achieved satisfactory registration outcomes. The main contributions of this paper are as follows:

1. We propose LPRnet, a novel method to capture robust features from irregular low overlap point clouds for registration, achieving superior performance on measured LiDAR and photogrammetric point cloud datasets.

2. We introduce a masked autoencoder and registration feature embedding module for registration feature extraction, corresponding to heterogeneous point cloud characteristics and rigid registration mechanism respectively.

3. A transformer-based global feature integration framework is designed to mitigate the impact of varying point cloud precision. By integrating this framework with an optimization algorithm, the proposed approach eliminates the reliance on ground truth for registration.

The rest of this paper is organized as follows. Section II introduces characteristic analysis of heterogeneous point clouds. Section III describes the self-supervised point cloud registration method in detail. Section IV introduces the experimental results and analysis of the proposed method on two datasets. Finally, the conclusions are drawn in Section V.

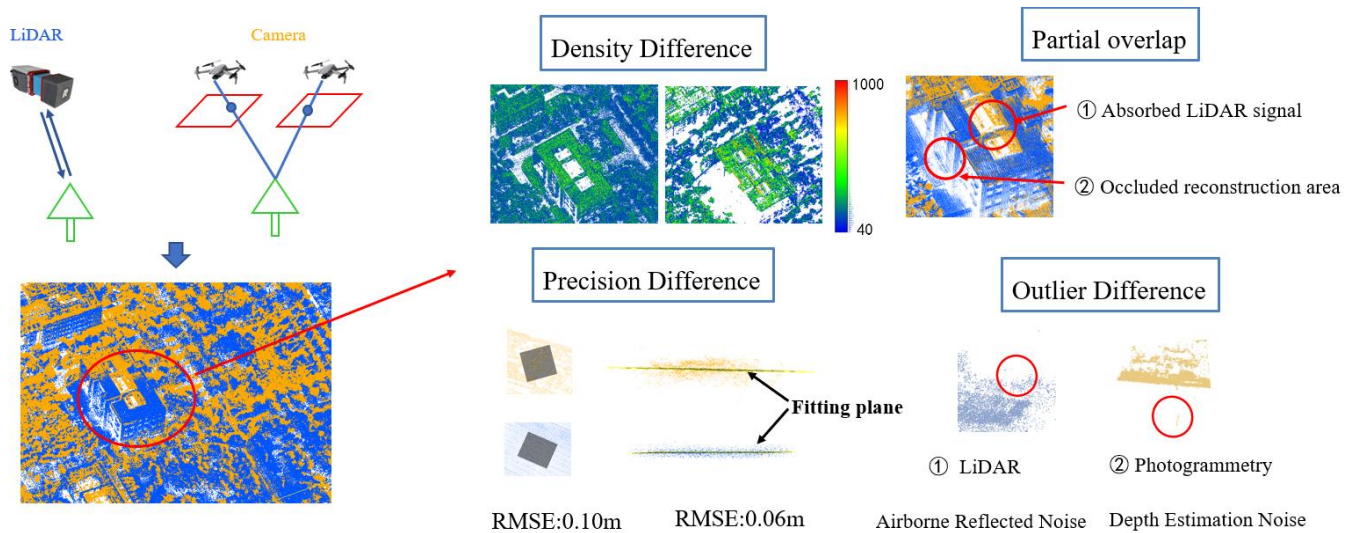

Fig. 1. The differences in point cloud characteristics between LiDAR and photogrammetry. Blue represents LiDAR point cloud, and yellow represents photogrammetric point cloud

## II. CHARACTERISTIC ANALYSIS OF HETEROGENEOUS POINT CLOUD DATA

This section introduces the principles of LiDAR and photogrammetric point cloud acquisition, providing a detailed analysis of their respective data characteristics and specific differences.

LiDAR systems mounted on UAVs emit laser pulses through a laser source, with the receiver capturing the reflected signals. By measuring the time difference between emission and reception, the system calculates the distance from the measurement point to the laser source, thereby determining the relative position of the point to the LiDAR. When combined with the drone's onboard GNSS/INS system, three-dimensional point cloud data with absolute geographic coordinates can be generated through point cloud processing algorithms. As a specialized active 3D measurement instrument, LiDAR produces point clouds with high accuracy, long detection range, and some degree of penetration capability. It is widely used in topographic mapping, forestry, and urban modeling [39, 40]. However, due to limitations in laser manufacturing processes, LiDARs typically operate within specific wavelength bands, such as 905 nm or 1550 nm, and are usually expensive. The point clouds generated by single-band LiDAR lack spectral and image texture information, while photogrammetric point clouds can effectively solve this limitation.

The photogrammetry method estimates the three-dimensional coordinates of the reconstructed points through multiple images with overlapping areas. Platforms equipped with a single optical camera or stereo cameras with multiple different angles collect images for photogrammetric point cloud generation. This method is cost-effective and commonly used in applications such as topographic mapping and heritage preservation [41]. The generated point clouds contain spectral information, but due to the nature of the generation process, they typically exhibit lower geographic coordinate accuracy and are more susceptible to occlusion. The point cloud is concentrated on the surface of the object and lacks penetration capability. Additionally, depth cameras and structured light cameras can also capture 3D point clouds; however, due to their shorter acquisition range, they are not suitable for use in remote sensing applications.

The differences in point cloud characteristics between LiDAR and photogrammetry are primarily evident in four aspects: density, partial overlap, precision, and outlier noise, as illustrated in Fig. 1.

1. Density distribution: The density of LiDAR point cloud is influenced by the laser emission frequency and the route design of the airborne platform, including path planning, altitude, and speed. Generally, increasing the density of LiDAR point clouds often comes at the cost of reduced operational efficiency. In contrast, the density of photogrammetric point cloud depends on the camera's resolution, the number of captured images, and the texture information of the observed scene. While photogrammetric point clouds typically exhibit higher density than LiDAR point clouds, they are often more unevenly distributed.

2. Partial overlap: As an active sensing instrument, LiDAR is significantly affected by the reflectivity of objects in the laser wavelength range. Commonly used laser wavelengths can markedly reduce the signal-to-noise ratio for materials like asphalt or darker surfaces, making surface point cloud generation challenging. In contrast, photogrammetric point cloud relies on at least two overlapping images for reconstruction, making occlusions caused by viewing angles the primary reason for missing reconstructed points. Insufficient image overlap can result in the loss of occluded areas, leading to incomplete point cloud. Additionally, LiDAR point clouds possess some penetration ability, allowing the capture of ground points beneath tree canopies, while photogrammetric point clouds are limited to the object surface and lack penetration capability.

3. Detection precision: Utilizing optical ranging technology, LiDAR typically achieves high geometric precision, with

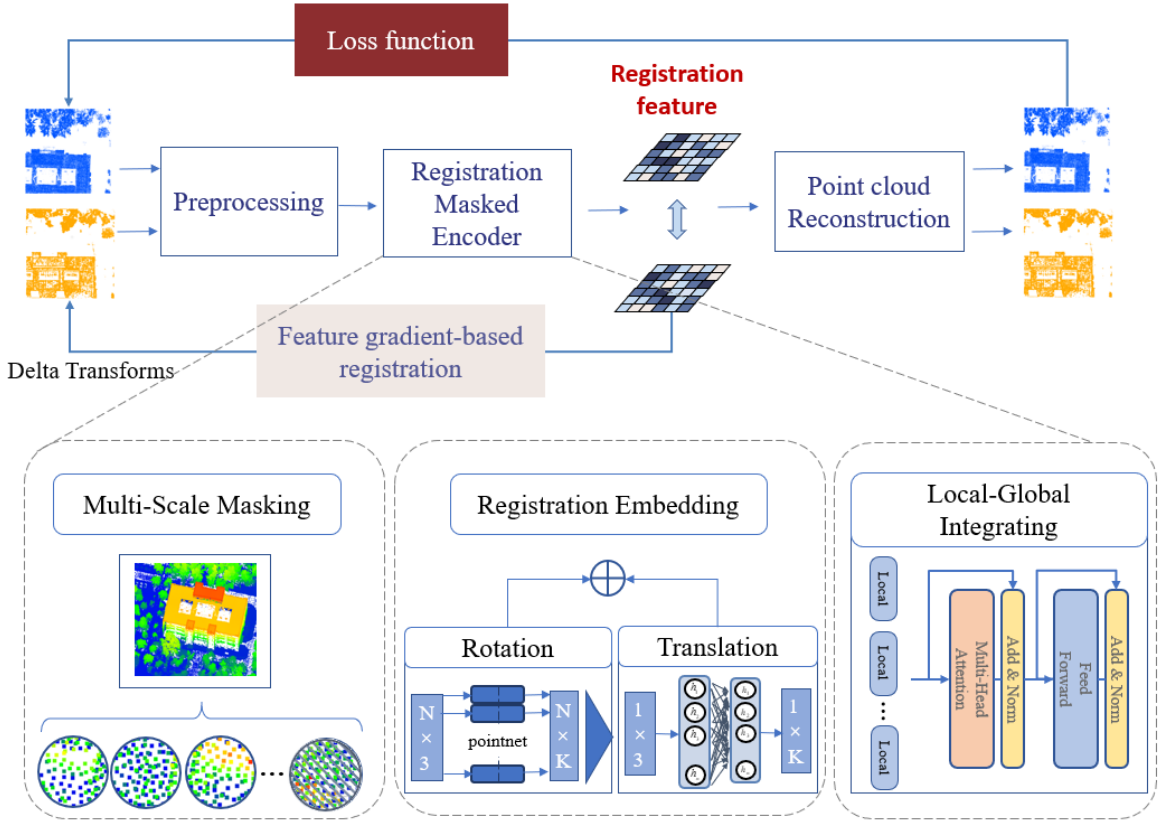

Fig. 2. The overall pipeline of self-supervised masked autoencoder registration network.

distance measurement precision usually at the centimeter level and effective ranges extending to several hundred meters. In contrast, the precision of photogrammetric point clouds is strongly influenced by image resolution. Based on empirical formulas, the planar precision of reconstructed models is approximately equal to the Ground sample distance (GSD), while the elevation precision is about three times the GSD.

4. Outlier noise: LiDAR outliers primarily arise from two factors: occlusion caused by birds or other flying objects and scattering or absorption of laser pulses due to environmental conditions. In photogrammetry, outliers occur primarily due to image noise, which leads to errors in depth estimation. There are obvious differences in the distribution characteristics of outliers between the two methods.

## III. SELF-SUPERVISED MASKED AUTOENCODER POINT CLOUD REGISTRATION METHOD

Based on the differences in point cloud distribution characteristics, the proposed point cloud registration framework bypasses direct point or feature-based registration, adopting an end-to-end approach that mitigates the effects of point cloud precision and density. In the network, we introduce a masked autoencoder of point clouds to extract features tailored to the partial overlap between heterogeneous point clouds. We combine local feature and position embedding with a transformer to obtain registration features. During training, a point cloud reconstruction loss function is used to extract robust features. In the registration phase, an optimization algorithm resolves the transformation relationship by following the gradient direction of heterogeneous point cloud features. The overall pipeline is illustrated in Fig. 2.

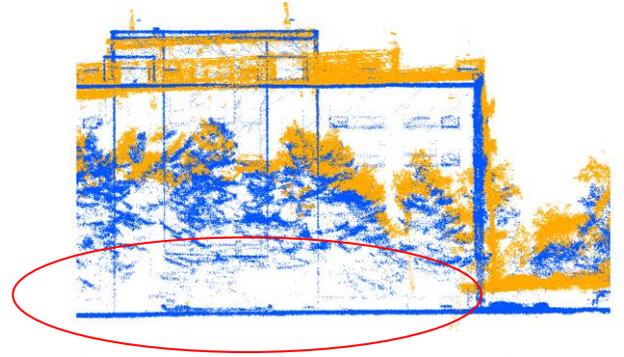

Fig. 3. Distribution of LiDAR point cloud and photogrammetric point cloud. Blue is the lidar point cloud and yellow is the photogrammetric point cloud.

*A. Point cloud generation and preprocessing*

To align heterogeneous point clouds in three-dimensional space, it is necessary to unify their spatial coordinate systems. For LiDAR point clouds, the absolute geographic coordinates of the points are computed using position and attitude data provided by GNSS/INS. Post-processing and filtering of GNSS/INS data with ground base station further enhances accuracy. For photogrammetric point cloud, ground control points collected by RTK are added during the reconstruction process to standardize the coordinates, ensuring that all point clouds are unified within the UTM coordinate system.

Due to the greater penetration of LiDAR point cloud compared to photogrammetric point cloud, their distribution of ground points differs significantly, as illustrated in Fig. 3. The additional ground points can introduce outlier noise during the registration process; therefore, the input point cloud undergoes pre-processing through ground point filtering to eliminate significant interference in calculating the registration relationship. While this filtering removes certain points, the retained ground points are included in the final fusion result to preserve data integrity. The filtering employs the Cloth Simulation Filtering method [42], which is applicable across various terrains and scenes, automatically adapting to different environments while maintaining high efficiency.

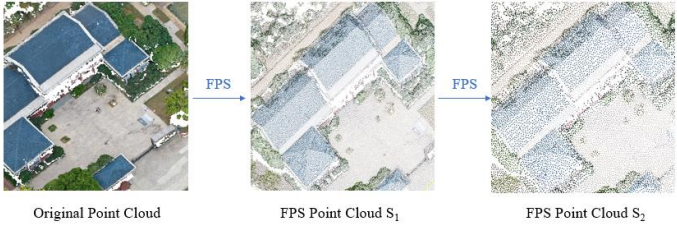

Fig. 4. Illustration of multi-scale FPS point cloud sampling results

*B. Masked Autoencoder Registration Network*

To cope with partially overlapping heterogeneous point clouds, we designed a point cloud registration network for feature extraction. This network primarily consists of three components: Multi-scale masking strategy, Registration feature embedding, and Encoder-Decoder with Prediction Head

1) **Multi-scale masking strategy**

The masking stage is designed to simulate the distributional differences between LiDAR and photogrammetric point clouds, enabling the encoder to extract more robust registration features. As concluded in Section II, the missing data in heterogeneous point clouds arises from various complex factors, leading to irregular low-overlap rates between the point clouds. To address this issue, a hybrid strategy combining multi-scale random masking and block masking is adopted during the masking stage.

Unlike masked autoencoders for images, point cloud data inherently lacks a fixed order. To define point cloud patches, Farthest Point Sampling (FPS) is employed to identify patch center. FPS preferentially selects points with the maximum mutual distance, which helps mitigate the effects of varying point cloud densities on the sampled point distribution. Observations of heterogeneous point cloud characteristics reveal that missing data is unrelated to point cloud density. To avoid the drawbacks of directly applying K-Nearest Neighbors (KNN) search, which might be sensitive to density variations, KNN is instead applied within multi-scale FPS-sampled point clouds. This approach effectively simulates missing data at different scales. An example of multi-scale FPS-sampled point clouds is illustrated in Fig. 4. This masking strategy ensures robustness against point cloud density variations and improves feature learning for irregular, low-overlap heterogeneous point clouds.

2) **Registration feature embedding**

In embedding stage, we focus on capturing features essential for registration rather than extracting high-level semantic features. Since the rigid transformation required for registration is determined by a combination of the rotation matrix and translation vector, we designed a rotation-translation feature embedding module to fulfill the requirement.

For visible point cloud patches, we adopt a PointNet-like structure for local feature embedding (FE). Compared with a simple MLP, the PointNet architecture is better suited for processing the unordered nature of point cloud data. In contrast to traditional PointNet or PPF-FoldNet approaches, our feature extraction network does not enforce rotational invariance. To achieve this, we removed the T-net component responsible for mapping invariance and introduced skip connections, enabling the network to capture low-level geometric details and rotational information from local point clouds. For translation information, position embedding (PE) is applied to the center of each point cloud patch, extracting local spatial information. Together, these two components form the embedding representation for each visible point cloud patch.

For the masked (invisible) regions, a shared-weight mask token is introduced to represent the missing parts, ensuring consistency in the embedding process. This combination effectively captures the local rotation-translation features while accommodating the unique characteristics of visible and masked regions.

3) **Encoder-Decoder with Prediction Head**

The encoder consists of a transformer architecture, which extracts contextual information from point cloud patches to obtain global features. The transformer architecture, inherently insensitive to input order, is particularly well-suited for point cloud data and masked autoencoder framework. Its self-attention mechanism effectively captures global feature correlations among local point clouds, even when points are spatially distant. By leveraging the masking strategy, the encoder focuses on extracting meaningful features from visible regions while balancing global and local characteristics.

In this stage, the FE and PE of visible tokens are input as shown in Equation 1.

$$F_n = T(FE + PE) \quad n = 1,2,...,N \quad (1)$$

Where $F_n$ are the local relational features, $T$ represents the encoder of the network.

The transformer block processes these inputs to produce encoding features F1 to FN of the point cloud. To enhance computational efficiency and prevent positional information leakage, only visible tokens are input during encoding.

In the decoder stage, a more lightweight structure is employed with fewer transformer blocks. Both visible and masked tokens are input into the decoder to generate the decoding features. Finally, a fully connected layer serves as the prediction head, reconstructing the point cloud.

4) **Loss functions**

A common loss function in point cloud registration is the difference between the network's output transformation and the ground truth of the registration transformation. However, in the absence of ground-truth transformation, our self-supervised registration framework trains network weights by reconstructing the target point cloud through a masking strategy. Consequently, the loss function employs the L2 chamfer distance between the input point cloud and the reconstructed point cloud to assess the reconstruction accuracy [33, 43], as described in Equation 2.

$$Loss = \frac{1}{|P_{pre}|} \sum_{a \in P_{pre}} \min_{b \in P_{masked}} \|a-b\|_2^2 + \frac{1}{|P_{masked}|} \sum_{b \in P_{masked}} \min_{a \in P_{pre}} \|a-b\|_2^2 \quad (2)$$

Where $P_{pre}$ represents the predicted point cloud, $P_{masked}$ represents the masked point cloud.

### C. Feature Gradient-based Registration Stage

In the registration stage, the proposed method integrates the reconstruction network with a nonlinear optimization algorithm to estimate the transformation matrix. This approach leverages the learned features from the reconstruction network to guide the optimization process, allowing for robust alignment of the source and target point cloud. The point cloud registration problem can be mathematically modeled as described in Equation 3.

$$\arg\min_{R,t} \|P_T - (R \cdot P_S + t)\|_2 \quad (3)$$

$P_S$ and $P_T$ represent the source and target point clouds, respectively. $R$ denotes the rotation matrix, and $t$ represents the translation vector. They form a rigid transformation relationship together.

Furthermore, employing a feature extraction network that is sensitive to the spatial relationships within the point cloud allows Equation 3 to be equivalent to Equation 4. Consequently, the point cloud registration problem can be reformulated as finding the transformation matrix that minimizes the objective in Equation 2.

$$\arg\min_{R,t} \|F(P_T) - F(R \cdot P_S + t)\|_2 \quad (4)$$

We denote the optimal rigid transformation by H, assuming that:

$$F(P_T) = F(H(P_S)) \quad (5)$$

To enhance computational efficiency, we employ an inverse compositional representation of Equation 5.

$$F(P_S) = F(H^{-1}(P_T)) \quad (6)$$

By expanding the above equation linearly, we obtain Equation (7).

$$F(P_S) = F(P_T) + \frac{\delta}{\delta \xi}[F(H^{-1} \cdot P_T)]\xi \quad (7)$$

According to Lie group and Lie algebra theory, $H$ can be transformed exponentially to simplify computation.

$$H^{-1} = \exp(-\sum_i \xi_i \mathbf{T}_i) \qquad \xi = (\xi_1, \xi_2, ..., \xi_6)^T \quad (8)$$

$\mathbf{T}_i$ represents the exponential mapping of the distortion parameter $\xi$.

To obtain the optimal transformation, calculating the Jacobian matrix of the feature extraction network with respect to the torsion parameters is required, but this process incurs significant computational cost.

$$\mathbf{J} = \frac{\partial}{\partial \xi}\left[F(H^{-1} \cdot P_T)\right] \quad (9)$$

In two-dimensional images, the classic Lucas-Kanade (LK) algorithm decomposes the Jacobian matrix using the chain rule. However, this approach cannot be directly applied to three-dimensional unordered point clouds. Inspired by PointNet-LK, the stochastic gradient method is employed to compute the Jacobian matrix $\mathbf{J}$.

$$\mathbf{J}_i = \frac{F(\exp(-t_i \mathbf{T}_i) \cdot P_T) - F(P_T)}{t_i} \quad (10)$$

where $t_i$ denotes a small perturbation in the distortion parameter within the stochastic gradient method. $\mathbf{J}_i$ represents the i-th column of the Jacobian matrix $\mathbf{J}$.

The Inverse Compositional Lucas-Kanade (IC-LK) algorithm utilizes the Jacobian matrix $\mathbf{J}$ calculated from the target point cloud, instead of recalculating it multiple times for the constantly changing source point cloud, significantly enhancing computational efficiency. Finally, the torsion transformation increment can be obtained as follows

$$\xi = \mathbf{J}^+ \left[F(P_S) - F(P_T)\right] \quad (11)$$

$\mathbf{J}^+$ is the generalized inverse matrix of $\mathbf{J}$.

The rigid transformation from the source point cloud to the target point cloud is determined by computing the Jacobian matrix. The optimal transformation is achieved through iterative updates, with transformation increments superimposed in each iteration.

## IV. EXPERIMENTS AND ANALYSIS

### A. Experimental dataset

We validate the proposed point cloud registration method using two large-scale remote sensing point cloud datasets and a simulated dataset. The experimental data were collected using two types of LiDAR, along with multispectral and optical cameras, to demonstrate the robustness of the registration method.

The LiDAR systems used include the RIEGL UAV mini3 and Velodyne VLP-16 Lite, with a comparison of their specifications provided in Table I. The cameras used in the experiments were the Micasense RedEdge dual-camera system and the Hasselblad L1D-20C optical camera, with their specifications compared in Table II.

1) **HIT campus**: This dataset was acquired using a RIEGL UAV mini3 LiDAR and a Micasense RedEdge MX dual multispectral camera. The LiDAR emitted pulses at 100,000 Hz from a flight altitude of 80 meters and at a flight speed of 8 m/s. The APX-15 served as the positioning and orientation system (POS) module for the LiDAR. The point cloud data was processed using specialized software RiPROCESS. A total of 480 images were captured by the multispectral camera, with an image overlap rate of 60%. Point cloud reconstruction followed the method outlined in [44]. The dataset is illustrated in Fig. 5 (a) and (b). The LiDAR point cloud consists of 20,917,231 points with an average point density of 190 points per square meter, while the multispectral photogrammetric point cloud consists of 6,895,831 points with an average density of 90 points per square meter.

TABLE I
THE MAIN PARAMETERS OF LiDARs

| Specification | RIEGL UAV mini3 | Velodyne VLP-16 |
|---|---|---|
| Range accuracy | ± 10 mm | ± 30 mm |
| Max Range | 270 m at 20% reflectivity | 100m |
| Pulse Frequency | 100,000–300,000 pulses/second | 300,000 pulses/second |
| Field of View | Horizontal 360° | Vertical 30°, Horizontal 360° |

TABLE II
THE MAIN PARAMETERS OF CAMERAS

| Specification | Rededge-MX dual | Hasselblad L1D-20C |
|---|---|---|
| Sensor resolution | 1280 × 960 px | 5472 × 3648 px |
| Field of view | 47.2° | 77° |
| Focal length | 5.5mm | 28mm |
| Ground Sample Distance at 120m | 8 cm/px | 3 cm/px |

2) **ZJK Mangrove**: A Velodyne VLP16 Lite LiDAR and a Hasselblad L1D-20C optical camera were used for data acquisition in this dataset. The LiDAR emitted pulses at a frequency of 100,000 Hz, with the system flying at an altitude of 80 meters and a speed of 5 m/s. The point cloud data was processed using LiDARtool. A total of 294 images were captured by the optical camera, with an image overlap of 60%. The reconstruction process was completed using Agisoft Metashape. The dataset is shown in Fig. 5 (c) and (d). The LiDAR point cloud contains 44,338,316 points with an average density of 280 points per square meter, while the optical photogrammetric point cloud contains 74,615,469 points with an average density of 600 points per square meter.

3) **Simulated dataset**: This dataset was processed using raw data from the RIEGL UAV mini3 LiDAR system, with simulated dataset generated by introducing varying levels of noise, occlusion, and rigid transformations [38]. Gaussian noise was added to the original point cloud data, with a mean of 0 and a standard deviation set to the accuracy of the photogrammetric point cloud, approximately 0.1m. The random occlusion rate was set to 20%-50%. The rotation range in the rigid transformation was set to 0°-30°, and the translation range was set to 0-2 meters. Ablation experiments of the proposed method are conducted by analyzing the registration performance of the method on this simulated dataset.

*B. Comparison Methods and Evaluation Metrics*

We compared the proposed method with several state-of-the-art point cloud registration techniques across multiple types, as well as representative methods in deep learning. These methods include traditional hand-crafted features like FPFH [21], the adaptive ICP-based approach FAST-ROBUST-ICP (Robust ICP) [18], cross-source point cloud registration methods such as the feature-learning approach VRHCF [31], and end-to-end deep learning methods like PointNetLK [36] and the advanced semi-supervised registration method FMR [38]. Due to the lack of ground truth in large-scale heterogeneous remote sensing point clouds, we used pre-trained parameters from each experimental method requiring supervised training when conducting comparisons.

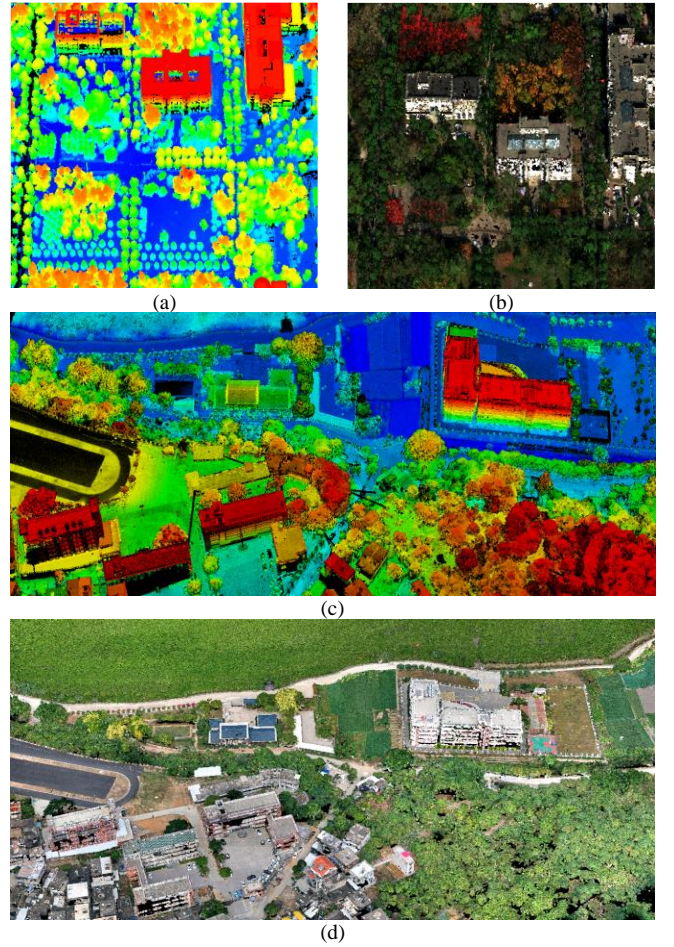

Fig. 5. Heterogeneous point cloud dataset (a) LiDAR point cloud of HIT campus. (b) Photogrammetric point cloud of HIT campus. (c) LiDAR point cloud of ZJK Mangrove Nature Research Center. (d) Photogrammetric point cloud of ZJK Mangrove Nature Research Center.

For the experimental evaluation metrics, the root mean square error (RMSE) of the nearest neighbor points between the point clouds was utilized, as the ground truth transformation matrix is not available in the measured dataset. The formula for this evaluation is presented below:

$$\text{RMSE} = \sqrt{\frac{1}{N}\sum_{i=1}^{N}\min_{j}\|p_i - q_j\|^2} \qquad (12)$$

where $p_i$ is the point in registered source point cloud, $q_j$ is the point in target source point cloud.

Due to factors such as partial overlap, differences in density, noise, and precision between point clouds, RMSE will not be zero even under optimal registration conditions. Therefore, in the context of heterogeneous point cloud registration, RMSE should be considered as one of the overall evaluation metrics, and it needs to be supplemented with evaluations based on DSM differences and visual results. To further reduce calculation errors arising from point cloud precision and partial overlap, the elevation difference of the digital surface model (DSM), commonly used in airborne remote sensing, was also employed as an additional evaluation metric.

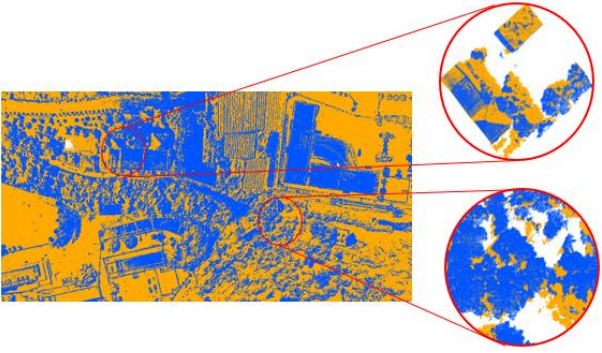

Fig. 6. The superimposed result of ZJK Mangrove Nature Research Center

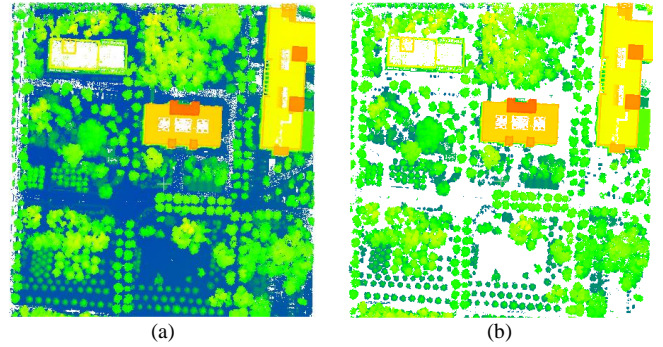

Fig. 7. The results of the HIT campus LiDAR point cloud before and after ground point filtering. (a) before filtering. (b) after ground point filtering

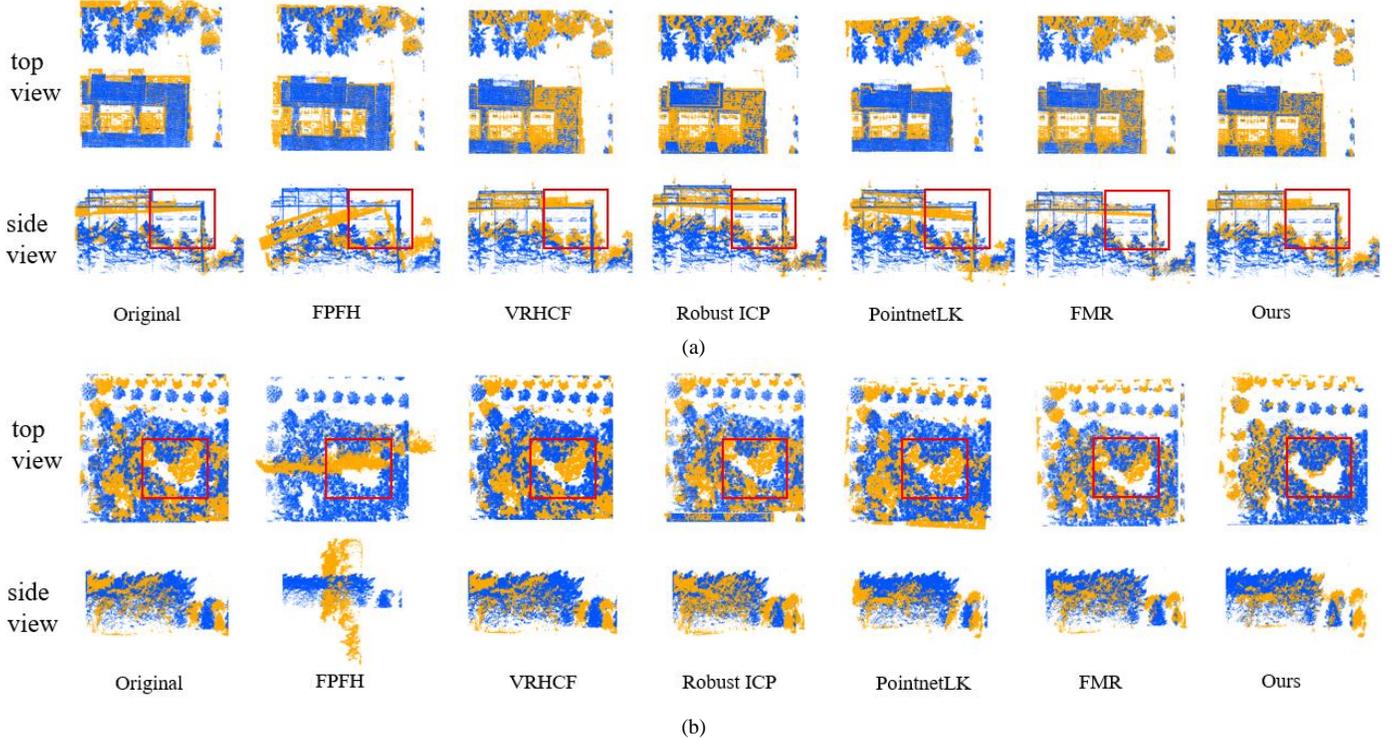

Fig. 8. Experimental results on local regions of point cloud registration. (a) Registration results of local building areas. (b) Registration results of local vegetation areas.

In the simulated data set, since the true value of the registration is known, we use root square mean error of transformation (RMSE-T) to evaluate the registration performance:

$$\text{RMSE-T} = \sqrt{\frac{1}{N}\sum_{i=1}^{N}||Test - Tgt||_F} \qquad (13)$$

Where $Test$ represents the estimated transformation and $Tgt$ represents the ground truth transformation. $||\cdot||_F$ is the Frobenius norm.

*C. Experimental results and analysis*

The proposed method is composed of three key stages: Point cloud generation and preprocessing, masked autoencoder training, and optimization of the registration transformation matrix. This section provides an analysis of the function and experimental results of each stage.

1) In the generation and preprocessing stage, the LiDAR data is processed using RIEGL specialized software, RIPROCESS, which accurately resolves the LiDAR point cloud. The coordinate information from the POS module is integrated during this process, and the coordinate system is set to the UTM-WGS 84 projection system. GNSS data from the camera is also incorporated to generate the 3D reconstructed point cloud, ensuring consistency in the point cloud's coordinate system. The superimposed result is shown in Fig. 6. While the point clouds largely occupy the same area, there is a noticeable spatial registration error.

After applying the CSF algorithm to filter out ground points, the number of lidar points decreases by 47%, and the number of reconstructed points decreases by 22%. This indicates that the proportion of ground points in the lidar point cloud is significantly higher than in the photogrammetric point cloud, and the ground points can be considered as the non-overlapping areas during the registration process. Fig. 7 shows the results of the HIT campus LiDAR point cloud before and after ground point filtering.

TABLE III
QUANTITATIVE EVALUATION RESULTS OF HIT CAMPUS REGISTRATION ACCURACY

| Methods | | Original | FPFH | VRHCF | Robust ICP | PointnetLK | FMR | Our |
|---|---|---|---|---|---|---|---|---|
| Region 1 | RMSE(m) | 0.76 | 0.94 | 0.62 | <u>0.40</u> | 0.52 | 0.47 | **0.34** |
| | DSM(m) | 3.71 | 5.54 | <u>3.42</u> | 3.51 | 3.45 | 3.65 | **2.77** |
| Region 2 | RMSE(m) | 0.98 | 1.08 | 1.06 | 1.02 | 1.30 | <u>0.72</u> | **0.60** |
| | DSM(m) | 3.84 | 4.72 | 4.56 | 3.88 | <u>3.74</u> | 3.80 | **3.64** |
| Region 3 | RMSE(m) | 0.66 | 1.15 | 1.74 | 0.48 | 0.58 | <u>0.42</u> | **0.38** |
| | DSM(m) | 3.08 | 4.60 | 11.07 | 3.01 | 2.95 | <u>2.68</u> | **2.41** |
| Region 4 | RMSE(m) | 0.50 | 0.70 | 0.43 | <u>0.35</u> | 0.64 | 0.48 | **0.34** |
| | DSM(m) | 2.11 | 5.43 | <u>1.74</u> | 1.76 | 2.95 | 1.81 | **1.72** |

TABLE IV
QUANTITATIVE EVALUATION RESULTS OF ZJK MANGROVE REGISTRATION ACCURACY

| Methods | | Original | FPFH | VRHCF | Robust ICP | PointnetLK | FMR | Our |
|---|---|---|---|---|---|---|---|---|
| Region 1 | RMSE(m) | 0.42 | 0.66 | 0.47 | **0.18** | 0.36 | 0.31 | <u>0.20</u> |
| | DSM(m) | 0.57 | 1.98 | 1.64 | <u>0.67</u> | 0.82 | 0.88 | **0.65** |
| Region 2 | RMSE(m) | 0.63 | 0.61 | 0.57 | <u>0.35</u> | 0.75 | 0.40 | **0.33** |
| | DSM(m) | 1.89 | 1.63 | 1.94 | 1.78 | 1.96 | <u>1.56</u> | **1.32** |
| Region 3 | RMSE(m) | 0.96 | 0.65 | 0.62 | <u>0.46</u> | 0.66 | 0.52 | **0.44** |
| | DSM(m) | 3.93 | 3.95 | 3.93 | 3.99 | 3.95 | <u>3.21</u> | **2.77** |
| Region 4 | RMSE(m) | 0.74 | 0.92 | <u>0.50</u> | 0.51 | 0.95 | 0.64 | **0.36** |
| | DSM(m) | 0.99 | 0.98 | 0.88 | 0.94 | 1.22 | <u>0.85</u> | **0.82** |

2) The point cloud masked autoencoder is designed to extract robust features from partially overlapping point clouds by leveraging the automatic masking mechanism. Compared with feature learning methods, this network does not rely on supervision or compute feature descriptors for individual points. Instead, it aims to capture registration-specific features that are sensitive to both position and rotation through point cloud reconstruction, as opposed to descriptors that are invariant to position and rotation.

For the mask generation, multi-scale FPS and KNN algorithms are employed to extract point cloud patches. The mask rate is determined based on the overlap rate of the heterogeneous point cloud data. To account for the overlap between patches, a random mask rate of 60% is applied. In the patch embedding stage, the position embeddings and local feature dimensions are set to 512. The backbone of the masked autoencoder consists of 8 Transformer blocks in the encoder and 4 Transformer blocks in the decoder. Each Transformer block has 384 hidden dimensions and 6 attention heads, with an MLP ratio of 4.

The experimental data were augmented using standard random rotations. The initial learning rate was set to 0.001, and a cosine learning rate decay was applied. Training was performed for 300 epochs using the AdamW optimizer, with a batch size of 4. The proposed method was implemented in PyTorch, and the experiments were conducted on NVIDIA Tesla V100.

3) We conducted experiments on two real outdoor scene heterogeneous point cloud datasets. The visualization of the experimental results is shown in Fig. 8. The quantitative evaluation results of the comparison methods are shown in Tables III and IV.

Considering the computational resource requirements as well as the registration performance for rigid transformations, we divided the point cloud into regions and processed them in blocks. The point cloud of each scene was divided into four regions, and the registration performance was evaluated for each region. Representative local point clouds from two datasets were selected for the visualization of registration results.

Fig. 8 (a) presents the registration results for the building area. It can be observed that the FPFH-based method, which relies on hand-crafted features, fails to successfully register the heterogeneous point clouds. This failure is expected, as FPFH is highly sensitive to noise and local missing data, making it inadequate for challenging registration tasks involving heterogeneous point clouds. Among optimization-based algorithms, ICP achieved relatively good global registration, but from the side view, it is evident that the relative rotational alignment of the heterogeneous point clouds in the local rooftop areas is not entirely accurate. Both deep learning approaches, PointNetLK and VRHCF, face two main challenges: dependence on ground truth for registration and the need for point cloud downsampling. The reliance on ground truth introduces feature domain discrepancies when registering outdoor scene-level data. Additionally, the requirement for downsampling when processing large-scale data results in loss of information. Through the design of the embedding module, our method demonstrates enhanced computational capacity to handle a larger volume of point clouds. The FMR method achieved registration results similar to Robust ICP, demonstrating robustness to point cloud density and noise. However, it did not achieve the best registration results compared to LPRnet.

TABLE V
ABLATION EXPERIMENTS OF LPRNET

| No | KNN | msFPS | PE | FE | RE | SMLP | TR | RMSE-T |
|---|---|---|---|---|---|---|---|---|
| 1 | √ |  | √ |  |  | √ |  | 0.34 |
| 2 | √ |  | √ |  |  |  | √ | 0.22 |
| 3 | √ |  |  | √ |  |  | √ | 0.18 |
| 4 | √ |  |  |  | √ |  | √ | 0.12 |
| 5 |  | √ |  |  | √ |  | √ | **0.09** |

Fig. 8 (b) shows the registration results for the vegetation area, a challenging dataset with both the irregularity of vegetation and partial data loss from heterogeneous sources such as missing trees in the photogrammetric point cloud. From the registration results, it is obvious that neither feature-based methods FPFH and VRHCF nor optimization-based methods Robust ICP were able to successfully register this area. The Robust ICP method even became trapped in a local optimum, yielding a transformation matrix with only minimal changes. The pointnetLK and FMR methods are also affected by the irregular low overlap characteristics of the point cloud and obtain incorrect registration matrices. End-to-end masked methods showed advantages in this scenario, as the proposed method was robust to areas with partial data loss in the heterogeneous point clouds and achieved effective registration for the entire region.

Quantitative evaluation results also demonstrate the superiority of the proposed method compared to other methods. It is important to note that, since there is no ground truth for evaluation, RMSE is subject to the influence of point cloud precision and partial overlap. For instance, while the iterative criterion of the ICP method aims to minimize the distance between point clouds, its RMSE is relatively low. However, as seen in the visualization results, the registration outcome is not optimal, and the DSM difference reveals projection errors in the elevation direction.

From Tables III and IV, across eight regions within two different scenarios, heterogeneous point clouds exhibit varying initial RMSE and DSM elevation differences. The proposed method consistently shows the best performance in most regions and demonstrates strong robustness to initial registration errors. Poor initial positioning often leads to incorrect registration results, such as ICP-based approaches that rely heavily on initial values. Supervised methods like PointNetLK and VRHCF show less stability on the tested datasets. FMR benefiting from its unsupervised nature, achieves second-best registration results in several regions of the HIT dataset. However, in ZJK mangrove scene, characterized by dense vegetation and partial point cloud loss, the registration accuracy is reduced. In contrast, the proposed LPRnet demonstrates robustness in registering photogrammetric and LiDAR point clouds.

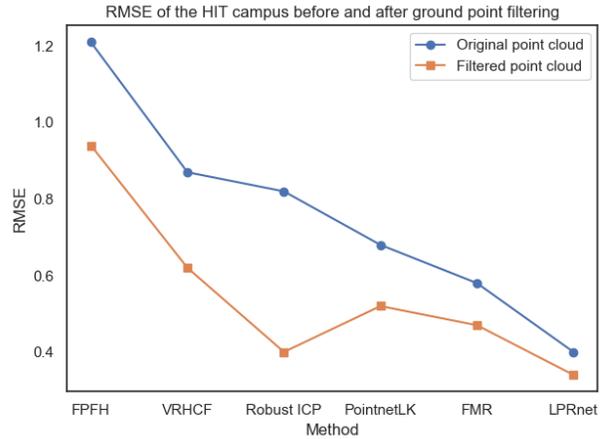

Fig. 9. Registration experiments of point cloud preprocessing

*D. Ablation experiments*

To evaluate the effectiveness of various modules in LPRnet, we first verified the impact of point cloud preprocessing and filtering. Subsequently, ablation experiments were conducted on the network modules, including masking module, embedding module, and integrating module. A complete model and four ablated models were proposed, with the experimental results presented in Table V.

**Preprocessing:** Registration experiments are conducted on the data before and after filtering, with the changes in RMSE results illustrated in Fig. 9. Non-overlapping ground points negatively impacted the registration performance, while the registration results improved significantly after filtering. Among the compared methods, both the FMR method and LPRnet were less affected by ground points, demonstrating the robustness to partially overlapping point clouds.

**Masking module:** The performance of the model without multi-scale masks on the simulated datasets is significantly degraded, as shown in (4) and (5) in Table V. This emphasizes the role of our multi-scale FPS (msFPS) module, which solves the KNN sampling density inconsistency problem and is effective for data registration of LiDAR and photogrammetric point clouds. This also indirectly shows that msFPS simulates the characteristics of heterogeneous point clouds more accurately than KNN sampling.

**Embedding module:** In the embedding module, we compared three models: position embedding (PE), feature embedding (FE), and the proposed registration feature embedding (RE). PE only utilizes MLP to embed point cloud tokens, FE applies PointNet for token embedding, and RE embeds rotational and translational features separately. As depicted in (2), (3) and (4), registration feature embedding is beneficial to the extraction of registration features and the improvement of registration accuracy.

**Integrating module**: In the integrating module, we compared the transformer architecture with a shared-weight MLP structure (SMLP) commonly used in PointNetLK and FMR. While both transformer and SMLP can aggregate global features and are well-suited for unordered point clouds, SMLP suffers from the loss of local features due to the pooling layers it employs. In contrast, the transformer architecture achieves superior integration of both local and global features. The transformer in the encoder shows an improvement in the integration of local and global features in the point cloud and generalization of the registration task.

## V. Conclusions

This paper presented a self-supervised registration method based on a masked autoencoder point cloud reconstruction network. Compared to existing registration methods, the proposed approach demonstrated significant advantages in heterogeneous point cloud registration, including robustness to variations in point cloud density, precision, noise, partial overlap, as well as its general applicability without requiring ground truth registration data. LPRnet extracted robust registration features through a multi-scale masking strategy, which enabled it to handle density differences and partial overlap of point clouds. The proposed method transformed point-to-point registration into a feature map-based registration process, reducing reliance on spatial precision of points. By integrating both global and local features, the influence of noise points is minimized. Experiments on both real and simulated datasets, with varying levels of noise and occlusion, further demonstrated the robustness of the method. Through experimental validation on two datasets and subsequent analysis of the results, it was concluded that the proposed method effectively registered heterogeneous point cloud data without supervised information, providing spatially consistent data for the fusion and utilization of point clouds.

## References

[1] Z. J. Li, B. Wu, Y. Li, et al. Fusion of aerial, MMS and backpack images and point clouds for optimized 3D mapping in urban areas, ISPRS Journal of Photogrammetry and Remote Sensing, 2023, 202: pp.463-478
[2] M. You, M. Guo, X. Lyu, et al. Learning a Locally Unified 3D Point Cloud for View Synthesis, IEEE transactions on image processing : a publication of the IEEE Signal Processing Society, 2023, 32: pp.5610-5622
[3] S. J. Zheng, W. Q. Liu, Y. Guo, et al. SR-Adv: Salient Region Adversarial Attacks on 3D Point Clouds for Autonomous Driving, IEEE Transactions on Intelligent Transportation Systems, 2024
[4] X. Zheng, X. Chen, X. Lu, et al. Unsupervised Change Detection by Cross-Resolution Difference Learning, IEEE Transactions on Geoscience and Remote Sensing, 2022, 60: pp.1-16
[5] M. Brell, K. Segl, L. Guanter, et al. 3D hyperspectral point cloud generation: Fusing airborne laser scanning and hyperspectral imaging sensors for improved object-based information extraction, ISPRS Journal of Photogrammetry and Remote Sensing, 2019, 149: pp.200-214
[6] Y. F. Gu, C. Wang, and X. Li. An Intensity-Independent Stereo Registration Method of Push-Broom Hyperspectral Scanner and LiDAR on UAV Platforms, IEEE Transactions on Geoscience and Remote Sensing, 2022, 60:
[7] X. Zheng, H. Sun, X. Lu, et al. Rotation-Invariant Attention Network for Hyperspectral Image Classification, IEEE Transactions on Image Processing, 2022, 31: pp.4251-4265
[8] P. Ghamisi, B. Rasti, N. Yokoya, et al. Multisource and Multitemporal Data Fusion in Remote Sensing: A comprehensive review of the state of the art, IEEE Geoscience and Remote Sensing Magazine, 2019, 7: pp.6-39
[9] Q. Wang, Y. Tan, and Z. Y. Mei. Computational Methods of Acquisition and Processing of 3D Point Cloud Data for Construction Applications, Archives of Computational Methods in Engineering, 2020, 27: pp.479-499
[10] W. Xiao, H. Cao, M. Tang, et al. 3D urban object change detection from aerial and terrestrial point clouds: A review, International Journal of Applied Earth Observation and Geoinformation, 2023, 118:
[11] N. Brightman, L. Fan, and Y. Zhao. Point cloud registration: a mini-review of current state, challenging issues and future directions, Aims Geosciences, 2023, 9: pp.68-85
[12] S. Monji-Azad, J. Hesser, and N. Löw. A review of non-rigid transformations and learning-based 3D point cloud registration methods, ISPRS Journal of Photogrammetry and Remote Sensing, 2023, 196: pp.58-72
[13] Y. F. Gu, Z. Xiao, and X. Li. A Spatial Alignment Method for UAV LiDAR Strip Adjustment in Nonurban Scenes, IEEE Transactions on Geoscience and Remote Sensing, 2023, 61:
[14] Z. Dong, F. X. Liang, B. S. Yang, et al. Registration of large-scale terrestrial laser scanner point clouds: A review and benchmark, ISPRS Journal of Photogrammetry and Remote Sensing, 2020, 163: pp.327-342
[15] X. Huang, G. Mei, and J. J. N. Zhang. Cross-source Point Cloud Registration: Challenges, Progress and Prospects, NEUROCOMPUTING, 2023, 548: pp.126383
[16] P. J. Besl, and N. D. McKay. A METHOD FOR REGISTRATION OF 3-D SHAPES, Sensor Fusion Conf Control Paradigms and Data Structures, 1991, 1611: pp.586-606
[17] J. L. Yang, H. D. Li, D. Campbell, et al. Go-ICP: A Globally Optimal Solution to 3D ICP Point-Set Registration, IEEE Transactions on Pattern Analysis and Machine Intelligence, 2016, 38: pp.2241-2254
[18] J. Y. Zhang, Y. X. Yao, and B. L. Deng. Fast and Robust Iterative Closest Point, IEEE Transactions on Pattern Analysis and Machine Intelligence, 2022, 44: pp.3450-3466
[19] G. F. Mei, F. B. Poiesi, C. Saltori, et al. Overlap-guided Gaussian Mixture Models for Point Cloud Registration, IEEE/CVF Winter Conference on Applications of Computer Vision, 2023 pp.4500-4509
[20] X. S. Huang, J. Zhang, Q. Wu, et al. A Coarse-to-Fine Algorithm for Matching and Registration in 3D Cross-Source Point Clouds, IEEE Transactions on Circuits and Systems for Video Technology, 2018, 28: pp.2965-2977
[21] R. B. Rusu, N. Blodow, M. Beetz, et al. Fast Point Feature Histograms (FPFH) for 3D Registration, IEEE International Conference on Robotics and Automation, 2009 pp.1848-1853
[22] R. B. Rusu, N. Blodow, Z. C. Marton, et al. Aligning Point Cloud Views using Persistent Feature Histograms, IEEE International Conference on Intelligent Robots and Systems, 2008 pp.3384-3391
[23] L. Chen, C. Z. Feng, Y. P. Ma, et al. A review of rigid point cloud registration based on deep learning, Frontiers in Neurorobotics, 2024, 17:
[24] J. J. Yu, F. H. Zhang, Z. Chen, et al. MSPR-Net: A Multi-Scale Features Based Point Cloud Registration Network, Remote Sensing, 2022, 14:
[25] Z. Y. Xu, Y. Zhang, J. T. Chen, et al. VMB Module for Low-Overlap Point Cloud Registration, IEEE Geoscience and Remote Sensing Letters, 2024, 21:
[26] C. Choy, J. Park, and V. J. I. Koltun. Fully Convolutional Geometric Features, IEEE International Conference on Computer Vision, 2019 pp.8957-8965
[27] H. Deng, T. Birdal, and S. Ilic. PPFNet: Global Context Aware Local Features for Robust 3D Point Matching, IEEE Conference on Computer Vision and Pattern Recognition, 2018 pp.195-205
[28] S. Ao, Q. Y. Hu, B. Yang, et al. SpinNet: Learning a General Surface Descriptor for 3D Point Cloud Registration, IEEE Conference on Computer Vision and Pattern Recognition, 2021 pp.11748-11757
[29] X. Y. Bai, Z. X. Luo, L. Zhou, et al. D3Feat: Joint Learning of Dense Detection and Description of 3D Local Features, IEEE Conference on Computer Vision and Pattern Recognition, 2020 pp.6358-6366
[30] F. Ghorbani, Y. C. Chen, M. Hollaus, et al. A Robust and Automatic Algorithm for TLS–ALS Point Cloud Registration in Forest Environments Based on Tree Locations, IEEE Journal of Selected Topics in Applied Earth Observations and Remote Sensing, 2024 pp.17
[31] G. Zhao, Z. Du, Z. Guo, et al. VRHCF: Cross-Source Point Cloud Registration via Voxel Representation and Hierarchical Correspondence Filtering, IEEE International Conference on Multimedia and Expo, 2024
[32] N. Ma, M. Wang, Y. Han, et al. FF-LOGO: Cross-Modality Point Cloud Registration with Feature Filtering and Local to Global Optimization, IEEE International Conference on Robotics and Automation, 2024

- [33] X. M. Yu, L. L. Tang, Y. M. Rao, et al. Point-BERT: Pre-training 3D Point Cloud Transformers with Masked Point Modeling, IEEE/CVF Conference on Computer Vision and Pattern Recognition, 2022 pp.19291-19300
- [34] Y. Pang, W. Wang, F. E. H. Tay, et al. Masked Autoencoders for Point Cloud Self-supervised Learning, European Conference on Computer Vision, 2022 pp.604-621
- [35] W. Lu, G. Wan, Y. Zhou, et al. DeepVCP: An End-to-End Deep Neural Network for Point Cloud Registration, IEEE/CVF International Conference on Computer Vision, 2019
- [36] Y. Aoki, H. Goforth, R. A. Srivatsan, et al. PointNetLK: Robust & Efficient Point Cloud Registration using PointNet, IEEE Conference on Computer Vision and Pattern Recognition, 2019
- [37] X. Li, J. K. Pontes, and S. Lucey. PointNetLK Revisited, IEEE/CVF Conference on Computer Vision and Pattern Recognition, 2020
- [38] X. Huang, G. Mei, and J. Zhang. Feature-metric registration: A fast semi-supervised approach for robust point cloud registration without correspondences, IEEE/CVF Conference on Computer Vision and Pattern Recognition, 2020 pp.11366-11374
- [39] C. Wang, Y. F. Gu, and X. Li. A Robust Multispectral Point Cloud Generation Method Based on 3-D Reconstruction From Multispectral Images, IEEE Transactions on Geoscience and Remote Sensing, 2023, 61:
- [40] M. M. Du, H. Y. Li, and A. Roshanianfard. Design and Experimental Study on an Innovative UAV-LiDAR Topographic Mapping System for Precision Land Levelling, Drones, 2022, 6:
- [41] K. Zainuddin, Z. Majid, M. F. M. Ariff, et al. 3D Modeling for Rock Art Documentation using Lightweight Multispectral Camera, International Workshop on 3D Virtual Reconstruction and Visualization of Complex Architectures, 2019, 42-2: pp.787-793
- [42] W. M. Zhang, J. B. Qi, P. Wan, et al. An Easy-to-Use Airborne LiDAR Data Filtering Method Based on Cloth Simulation, Remote Sensing, 2016, 8:
- [43] Y. Pang, W. Wang, F. E. H. Tay, et al. Masked Autoencoders for Point Cloud Self-supervised Learning, European Conference on Computer Vision, 2022, 13662: pp.604-621
- [44] C. Wang, X. Li, Y. F. Gu, et al. An adaptive 3D reconstruction method for asymmetric dual-angle multispectral stereo imaging system on UAV platform, Science China-Information Sciences, 2024, 67:

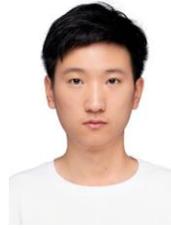

**Chen Wang** received the B.E. degree in electronics and information engineering from the Harbin Institute of Technology, Harbin, China, in 2019, where he is currently pursuing the Ph.D. degree in information and communication engineering. His current research interests include hyperspectral images and LiDAR data processing and its application in remote sensing.

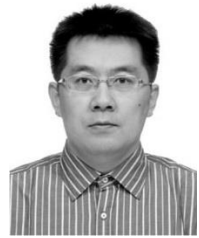

**Yanfeng Gu** (M'06-SM'16) received the Ph.D. degree in information and communication engineering from Harbin Institute of Technology, Harbin, China, in 2005. He joined as a Lecture with the School of Electronics and Information Engineering, Harbin Institute of Technology (HIT). He was appointed as Associate Professor at the same institute in 2006; meanwhile, he was enrolled in first Outstanding Young Teacher Training Program of HIT. From 2011 to 2012, he was a Visiting Scholar with the Department of Electrical Engineering and Computer Science, University of California, Berkeley, CA, USA. He is currently a Professor with the Department of Information Engineering, HIT, Harbin, China. He has published more than 100 peer-reviewed papers, four book chapters, and he is the inventor or coinventor of 20 patents. His research interests include space intelligent remote sensing and information processing, multimodal hyperspectral remote sensing, spaceborne time-series image processing.

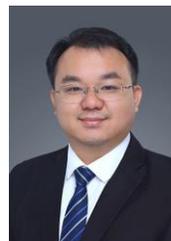

**Xian Li** (Member, IEEE) received the Ph.D. degree in instrument science and technology from the Harbin Institute of Technology (HIT), Harbin, China, in 2021. From 2018 to 2020, he was a Doctoral Researcher with the Department of Telecommunications and Information Processing, Ghent University, Ghent, Belgium, supported by the China Scholarship Council. He is currently an Assistant Professor with the School of Electronics and Information Engineering, HIT. His research interests include deep learning, multi-source remote sensing, and data processing.